\documentclass[a4paper,conference]{IEEEtran}
\usepackage{cite}

\ifCLASSINFOpdf
\else
\fi
\usepackage{nicefrac}  
\usepackage{xspace}
\usepackage{amsmath,amssymb,bm,stmaryrd}
\DeclareMathSymbol{\shortminus}{\mathbin}{AMSa}{"39}
\usepackage{enumitem}
\usepackage{flushend}

\usepackage{changepage}

\usepackage[utf8]{inputenc} 
\usepackage[T1]{fontenc}    
\usepackage{hyperref}       
\usepackage{url}            
\usepackage{booktabs}       
\usepackage{amsfonts}       
\usepackage{nicefrac}       
\usepackage{microtype}      
\usepackage{graphicx}
\usepackage{dblfloatfix}
\usepackage{algorithmic}
\usepackage{comment}
\usepackage{footnote}
\usepackage[ruled,vlined]{algorithm2e}
\usepackage{caption} 
\usepackage{sidecap}
\usepackage{bbm}
\usepackage{xcolor}
\usepackage{verbatimbox}    

\newcommand{\bS}{\mathbf{S}}
\newcommand{\bM}{\mathbf{M}}
\newcommand{\bp}{\mathbf{h}}

\newcommand{\bz}{\mathbf{z}}
\newcommand{\bem}{\mathbf{m}}
\newcommand{\past}{\text{p}}
\newcommand{\fut}{\text{f}}
\newcommand{\name}{DIVA}

\def\eg{\emph{e.g.}}

\def\etal{\emph{et al.}}

\newcommand{\laura}[1]{\textcolor{black}{#1}}

\begin{document}
%
\title{Diverse Probabilistic Trajectory Forecasting\\with Admissibility Constraints}



%

\author{\IEEEauthorblockN{Laura Calem\IEEEauthorrefmark{1}\IEEEauthorrefmark{2},
Hedi Ben-Younes\IEEEauthorrefmark{2},
Patrick Pérez\IEEEauthorrefmark{2} and
Nicolas Thome\IEEEauthorrefmark{1}}
\IEEEauthorblockA{\IEEEauthorrefmark{1}Conservatoire National des Arts et Métiers (CNAM), Paris, France}
\IEEEauthorblockA{\IEEEauthorrefmark{2}Valeo.ai, Paris, France\\
Email: {firstname.lastname@lecnam.net, firstname.lastname@valeo.com}
}}


\maketitle

\begin{abstract}
  Predicting multiple trajectories for road users is important for automated driving systems: ego-vehicle motion planning indeed requires a clear view of the possible motions of the surrounding agents.
  However, the generative models used for multiple-trajectory forecasting suffer from a lack of diversity in their proposals. 
  To avoid this form of collapse, we propose a novel method for structured prediction of diverse trajectories. 
  To this end, we complement an underlying pretrained generative model with a diversity component, based on a determinantal point process (DPP). 
  We balance and structure this diversity with the inclusion of knowledge-based quality constraints, independent from the underlying generative model.
  We combine these two novel components with a gating operation, ensuring that the predictions are both diverse and within the drivable area.
  We demonstrate on the nuScenes driving dataset the relevance of our compound approach, which yields significant improvements in the diversity and the quality of the generated trajectories.

\end{abstract}

%
\IEEEpeerreviewmaketitle

\section{Introduction}

In trajectory forecasting, future prediction is inherently stochastic since the human or automated driver has only access to very partial information about other road users' intents. It is also often multi-modal, since several admissible, yet very different driving actions can be taken at any instant by each agent. Intuitively, ignoring part of these possible future trajectories can hinder an autonomous or assisted driving system. It has been shown in \cite{cui2021lookout} that a diverse future generation improves performance in planning tasks, especially mitigating the issue of conservative driving in autonomous vehicles. For these reasons, multiple-output forecasting models have emerged. The main challenge is to predict a limited number of future trajectories that capture well the available driving options for the near future. A crucial aspect is thus to control the diversity of the proposed trajectory set.

Recent works have built on generative autoencoders to sample multiple future trajectories \cite{lee2017desire,park2020diverse,salzmann2020trajectron++}. However, the output distribution that such models provide sticks by construction to the one in the training data, which is mostly unimodal if real driving recordings are used: only a single future exists for a given past trajectory. At a higher level, some types of trajectories, such as turning rather than driving straight at an intersection, are severely under-represented.

\begin{figure*}[h!]
    \begin{center}
    \includegraphics[width=0.9\textwidth]{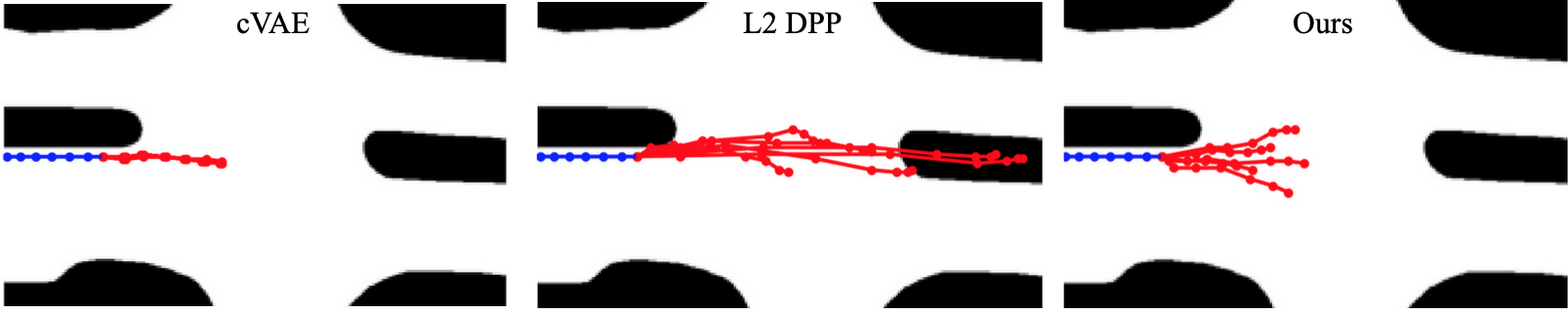}
    \end{center}
    \vspace{0pt}
    \caption[the caption]{\textbf{Effect of different methods on diversity}. Given a vehicle's known past trajectory (blue) and the road layout (black and white map), multiple futures are predicted (red). (Left) In real datasets, a single future trajectory is available in training, making a standard generative model such as a conditional VAE (cVAE) unable to sample admissible options far away from the supervision. 
    (Middle) Although Determinantal Point Processes (DPPs)~\cite{kulesza2012determinantal} are appealing for sampling diverse predictions, 
    using a standard $\ell_2$ kernel as in \cite{yuan2019diverse} induces mostly \textit{longitudinal} variations and may overshoot in non-drivable areas. (Right) In the proposed method, DIVA, the designed DPP kernel also considers the \textit{lateral} deviation at destination between two trajectories, and explicitly penalizes predictions outside the drivable area when training the diversity model. Consequently, DIVA samples driving options that are diverse, including steering-wise, and admissible.
    }
    
    \label{fig:intro}
\end{figure*}

Many generative models used for trajectory prediction \cite{lee2017desire, chai2019multipath, salzmann2020trajectron++}, \eg, based on generative adversarial networks (GANs)~\cite{goodfellow2014generative} or variational auto-encoders (VAEs)~\cite{razavi2018preventing, lucas2019understanding}, have no explicit control on the diversity beyond the one of the data distribution. Therefore, the dominant mode will be sampled every time, a problem exacerbated in our context where we aim at sampling a few trajectories only that summarize well the possible futures. 
This observation, illustrated in \autoref{fig:intro}, motivates our approach for designing a probabilistic model based on a more structured diversity. 

In this paper, we introduce a new method of DIVerse trajectory prediction with Admissibility constraints (DIVA) for probabilistic forecasting of road users. In particular, our approach allows the sampling of the main relevant modes of the future trajectory distribution, 
as illustrated in \autoref{fig:intro}. 
To achieve this goal, 
our contributions are:
\begin{itemize}[leftmargin=6pt,topsep=2pt,itemsep=0pt,parsep=0pt,partopsep=0pt]  
  \item We introduce a diversity sampling function (DSF) 
  based on a DPP~\cite{kulesza2012determinantal}. The diversity is explicitly controlled
through the definition of the DPP kernel. In particular, we introduce a new kernel adapted to the task at hand, which enforces trajectories' end-points to be far away in the \textit{lateral} direction (amounting to steering diversity) rather than in the \textit{longitudinal} one (amounting to speeding diversity). 
  \item We also control the ``quality'' of the sampled trajectories via a loss that penalizes violations of the driving area's topology. 
  We learn quality and diversity-based latent codes which we merge with a gating fusion mechanism. This enables the quality loss to filter out irrelevant trajectories predicted outside of the drivable area.   
  \item We evaluate the performance of our system on a real-world dataset (nuScenes \cite{nuscenes2019}) with a broad selection of metrics, demonstrating that trajectories that are both diverse and admissible are well produced. 
\end{itemize}

\section{Related work}

\noindent\textbf{Diversity.~} A growing body of research \cite{chai2019multipath, zhang2013dynamic, zhao2019multi, weng2020end, robicquet2016learning}
involves predicting a distribution of future trajectories rather than a univocal future. Many of these methods build upon an encoder-decoder architecture with sampling in the latent space, either with a traditional cVAE \cite{lee2017desire} or with more elaborated techniques \cite{alahi2016social, park2020diverse, salzmann2020trajectron++}. 
Ramashinghe \etal \cite{ramasinghe2020conditional} provide a mechanism for modeling the latent space as a continuous multimodal space, but assume that a distribution of admissible ground truths for each training example is available. This is often not the case in real-world driving datasets.

Several strategies have been applied to overcome this limitation.
In MTP \cite{mtp2019}, a multi-output architecture is proposed, trained to encourage each mode to specialize for a distinct behavior. 
In recent work \cite{lapred2021cvpr}, the lane information is used as a prior for semantic behavior decision, thus providing feasible and diverse trajectory forecasts.
Park \etal \cite{park2020diverse} use a normalizing flow \cite{rezende2015flow} decoder, and approximate the true distribution of future trajectories using the whole drivable area instead of the single ground truth, which encourages sample diversity.
CoverNet \cite{phan2020covernet} tackles the issue of diversity by predicting trajectories as distinct classes, where the set of possible categories is chosen to maximize the coverage on a training set.
Another line of approaches uses DPPs to increase the diversity in the set of predicted trajectories. DPPs, introduced in \cite{macchi1975coincidence} in the context of particle physics, are probabilistic models which recently gained the attention of the machine learning community \cite{kulesza2012determinantal,mariet2019dppnet,robinson2019flexible,celis2016complexity}. 
They have been explored for various applications such as video subset selection \cite{gong2014diverse}, document summarization \cite{hong2014improving}, or time series forecasting \cite{guen2020stripe}. 
GDPP \cite{elfeki2019gdpp} provides an interesting way to build the DPP kernel by matching the true diversity of the data. However, this method requires access to the ground-truth distribution of the data, which is not available in real-world driving datasets.
In the context of trajectory forecasting, DPPs have been used with cVAEs in \cite{yuan2019diverse} and with Graph Neural Networks in \cite{weng2021ptp}. In our work, we also use a DPP to improve the diversity of the predicted trajectories. We depart from these previous works by incorporating scene information in the DSF, which guides the sampling towards more admissible regions.

\smallskip\noindent\textbf{Admissibility.~} Several works explore using physical constraints to guide trajectory generation. 
In Neural Motion Planner \cite{motionplanner2019}, candidate trajectories are sampled in the space of \textit{clothoids}, which ensures that they are dynamically feasible. 
CoverNet \cite{phan2020covernet} generates a set of possible future trajectories by integrating the dynamic state of the vehicle.
Park \etal \cite{park2020diverse} generate physically-admissible trajectories by setting a low acceleration prior on the predictions.
While having no explicit control for admissibility, Salzmann \etal \cite{salzmann2020trajectron++} constrain the outputs to be admissible under the vehicle's current dynamic state. Our work differs from these works as we define admissibility with layout constraints in addition to dynamic feasibility. 


\section{Proposed Approach}

 We detail here the DIVA model for diverse trajectory prediction with admissibility constraints. DIVA builds upon a generative model to construct a latent space from which to sample codes representing future trajectories (\autoref{sec:cvae-formulation}). We then describe in \autoref{sec:model} the proposed method for introducing a structured diversity via a DPP kernel, while controlling the quality of the forecast with respect to the drivable area. 
 

\subsection{Problem formulation}
\label{sec:cvae-formulation}

Given the $T_{\past}$ past (and current) 2D positions of an agent and a ``map'' of its current environment, the multi-output forecasting task amounts to predicting $N$ possible trajectories over the $T_{\fut}$ future instants. Denoting $\bS = (\bS_{\past},\bS_{\fut})\in \mathbb{R}^{(T_{\past}+T_{\fut})\times 2}$ the agent's trajectory over the whole time interval and $\bM\in\mathbb{R}^{H \times W\times3}$ the environment map centered on agent's current position $\bS_{\past}(T_{\past})$ (using an RGB encoding of all static and dynamic elements in the scene, see example in \autoref{fig:archi} and detail in Supplementary A), the forecasting model is trained on example pairs $(\bS,\bM)$. At runtime,
it must predict for each agent in the scene $N$ trajectory samples, $\hat{\bS}_{\fut}^{(n)}, n=1\cdots N$, given $(\bS_{\past},\bM)$.     
Following \cite{park2020diverse}, the temporal horizons in our experiments are set to $T_{\past} = 12$ and $T_{\fut} =6$, which 
amounts to 6 and 3 seconds respectively at 2Hz, and the number of predictions is $N = 12$.




While our method is agnostic to the specific architecture of the underlying generative model, we chose for our experiments a simple conditional variational autoencoder (cVAE), as done in \cite{lee2017desire} for trajectory prediction, 
which we adapt to suit our specific needs, as explained next. 

\smallskip\noindent\textbf{Encoding.~} At a given instant and for a given agent in the scene, the encoding block takes $(\bS_{\past},\bM)$ as input. The past trajectory is encoded by a gated recurrent unit (GRU) network \cite{cho2014gru}, as
$\bp = \text{GRU}(\bS_{\past})$, where $\bp\in\mathbb{R}^{d_h}$ is the last hidden state of the recurrent network. The map of the agent's environment is processed by a convolutional neural network to produce an embedding $\bem = \text{CNN}(\bM)$ used as local physical constraints.


\smallskip\noindent\textbf{Sampling and decoding.~} Both embeddings $\bem$ and $\bp$ are concatenated and used to predict the parameters $\mu$ and $\sigma$ of the Gaussian distribution over latent codes $\bz\in\mathbb{R}^{d_z}$. A sampled latent code is then concatenated with $\bem$ and $\bp$ to produce the initialization for the hidden units of the decoder recurrent network. Finally, the output of this RNN decoder is passed through a series of fully-connected layers to produce the final trajectory $\hat{\bS}_{\fut}$. In effect, $N$ latent codes are sampled for a given $(\bem,\bp)$, yielding $N$ distinct future trajectories.\\

\smallskip\noindent\textbf{Training the generative model.~} To train the underlying generative model, we use the VAE loss introduced in \cite{kingmawelling2014}, adapted to include both inputs $\bS_{\past}$ and $\bM$ and to reflect the predictive nature of the task rather than an autoencoding one:
\begin{multline}
    L_{\text{cvae}}(\phi,\theta) = \mathbb{E}_{q_{\phi}(\bz | \mathbf{S}_{\past}, \bM)} [ \log p_{\theta}(\hat{\mathbf{S}}_{\fut} | \bz, \mathbf{S}_{\past}, \bM)]\\ 
    - \text{KL}(q_{\phi}(\bz | \mathbf{S}_{\past}, \bM) \| p(\bz)),
    \label{eq:vae_loss}
\end{multline}
where $\phi$ and $\theta$ are the parameters of the encoder and decoder respectively. 
The first term is the likelihood of the predicted trajectory and can be seen as a reconstruction quality term; the second term is the Kullback-Leibler divergence between the learned latent distribution $q_{\phi}$ and a prior $p(\bz)$, generally chosen to be Gaussian \cite{higgins2016beta, lee2017desire} for ease of sampling from this prior. 
%
A generative model alone usually suffers from mode collapse, as no incentive is provided to produce diverse samples. In that case, the trajectories generated by the model concentrate around the main mode from the underlying trajectory distribution, as illustrated in \autoref{fig:qu}.

\begin{figure*}[h!]
    \centering
    \includegraphics[width=0.95\textwidth]{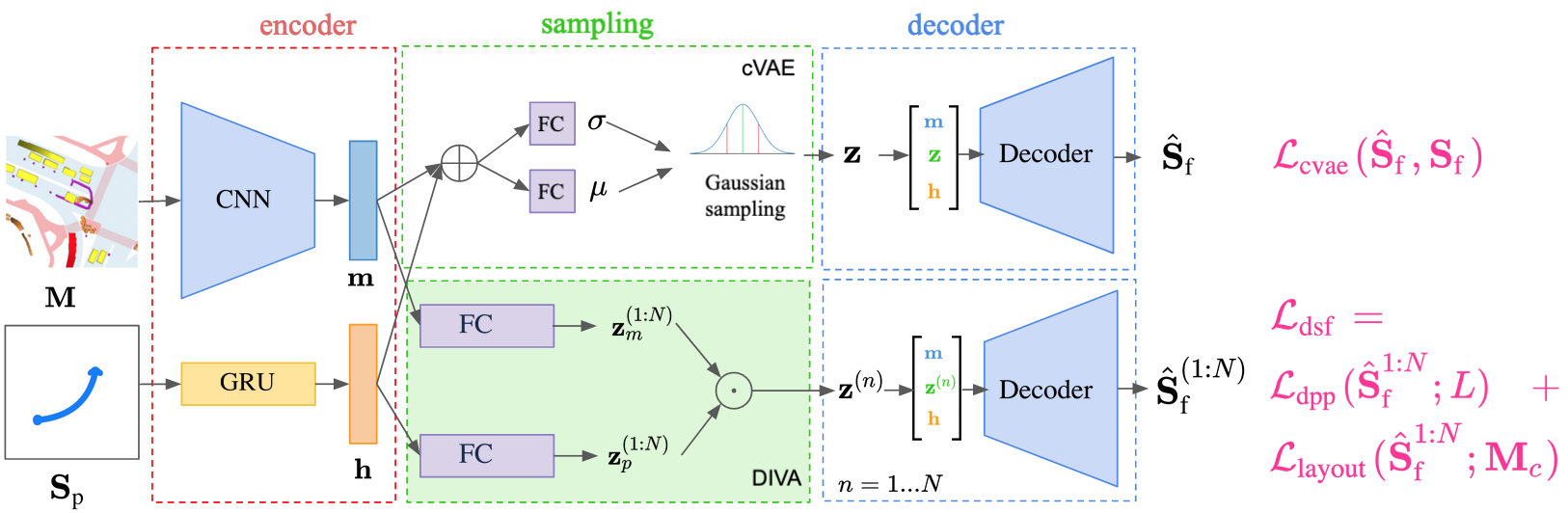}
    \vspace{0pt}
    \caption{\textbf{General architecture of the proposed trajectory prediction method in DIVA}. The upper part of the figure describes the underlying generative model, here a cVAE adapted to include layout information $\bM$. The lower part of the figure shows the proposed diversity sampling function that replaces the sampling part. $\bigoplus$ and $\bigodot$ denote concatenation and element-wise product, respectively.}
    \label{fig:archi}
\end{figure*}

\subsection{Structured diversity with physical constraints} \label{sec:model}


Given a trained generative model, we propose to replace the sequential random sampling from the prior $p(\bz)$ with a diversity sampling function (DSF) trained to predict multiple $\hat{\mathbf{S}}_{\fut}$'s jointly. 
As illustrated in the lower part of \autoref{fig:archi}, the DSF is implemented as a small two-branch feed-forward neural network. In contrast with the generative model sampling, where the $N$ latent codes are sampled independently in $\mathbb{R}^{d_z}$, the network is designed to output all the latent codes at once, producing an output in $\mathbb{R}^{N \times d_z}$.


In order to structure the diversity of the proposed trajectories, we split the DSF between diversity and quality, with each branch controlling a partial latent code. The diversity branch takes the representation of the past trajectory, $\bp$, and produces $N$ partial latent codes $\bz_p^{(n)}$, whereas the quality branch takes the map representation $\bem$ and gives $N$ partial latent codes $\bz_m^{(n)}$. The two associated partial codes are then combined using an element-wise product to produce a final latent code $\bz^{(n)}$. Through this gating mechanism, the map-specific constraints are imposed to the diverse set of trajectories.
The corresponding training loss,
\begin{equation}
    \mathcal{L}_{\text{dsf}} = \lambda \mathcal{L}_{\text{dpp}} + (1 - \lambda) \mathcal{L}_{\text{layout}},
    \label{eq:dsf_loss}
\end{equation}
is comprised of two terms. The first term, $\mathcal{L}_{\text{dpp}}$, favors the diversity  through an adapted DPP kernel and the second one, $\mathcal{L}_{\text{layout}}$, injects the quality constraints; $\lambda \in (0,1)$ is a parameter controlling the tradeoff between the two losses, as discussed in greater detail in \autoref{analysis}. Next, we detail these two loss terms.

\subsubsection{\textbf{Diversity with a DPP kernel}}

In the following section, we first provide some background on Determinantal Point Processes (DPPs), summarizing from \cite{kulesza2012determinantal} and \cite{yuan2019diverse}, in order to give enough context for our proposed method. DPPs are probabilistic set models that allow for an explicit handling of negative correlations among sets' elements. DPPs were first used in the context of particle physics for their ability to model the repulsion between particles: contrary to sampling a uniform distribution in space, which results in some level of clumping, sampling a DPP results in a more uniform spread.  

Formally, given a countable ground set $Y$ of ``items'', a DPP is a distribution over the power set of $Y$, giving the probability to draw any part of $Y$. It thus defines a random set, $\mathbf{A}$. We focus here on the class of DPPs that are defined through a positive semi-definite kernel $L$ as follows: for any finite subset $B$ of $Y$, $\mathbb{P}[\mathbf{A} =  B] \propto \text{det}(L_B)$, where $L_B$ is the matrix defined by $L$ over $B$ and the normalization constant has a closed form. 
It can also be shown that the probability $\mathbb{P}[\mathbf{A}\supset B]$ that the random set includes $B$ 
is exactly $\text{det}(K_B)$, where $K:= (L+\text{Id})^{-1}L$. In particular, $\mathbb{P}[\mathbf{A}\supset \{a\}] = K(a,a)$ for any $a\in Y$ and $\mathbb{P}[\mathbf{A}\supset \{a,b\}] = K(a,a)K(b,b)-K(a,b)^2$ for any pair. The latter provides insight into the repulsive behavior captured by the DPP: the more similar two items according to $K$ (and to $L$, as $K$ derives from $L$ through a rescaling of $L$'s eigenvalues), the more unlikely they are to be jointly included in $\mathbf{A}$.



Building upon this base DPP definition, we now explain how DPPs are integrated in our context. The goal is to produce a maximally diverse set 
of $N$ future trajectories. To this end, we follow \cite{yuan2019diverse} and define the ground set $Y$ as the finite set of the $N$ predicted trajectories. 
Intuitively, the overall diversity defined by $L$ over $Y$ reflects into the expected cardinality of the associated DPP. As this expectation reads 
$\mathbb{E}(|\mathbf{A}|) = \text{trace}[\text{Id} - (L_Y+\text{Id})^{-1}]$, see \cite{kulesza2012determinantal}, 
the expression in the r.h.s. can be used to define the diversity loss for the DSF. This yields:
\begin{equation}
    \mathcal{L}_{\text{dpp}}\big(\hat{\bS}_{\fut}^{(1:N)};L\big) = - \text{trace}\big[\text{Id} - (L_Y+\text{Id})^{-1}\big],
    \label{eq:dpploss}
\end{equation}
where $Y = \{\hat{\bS}_{\fut}^{(1)},\ldots,\hat{\bS}_{\fut}^{(N)}\}$ and $L$ is a kernel to be defined on trajectories. 
Given two future trajectories $\hat{\bS}^{(i)}_{\fut}$ and $\hat{\bS}^{(j)}_{\fut}$ predicted from a same past and present, the trajectory kernel can be simply defined as a spherical Gaussian kernel. This, however, proves insufficient to promote \textit{directional} diversity among the generated trajectories. Hence, we also include in the kernel the angular deviation between the final points of the two trajectories:
\begin{equation}
    L\big(\hat{\bS}^{(i)}_{\fut},\hat{\bS}^{(j)}_{\fut}\big) = \exp -\alpha \big(\theta_{ij} + \| \hat{\bS}^{(i)}_{\fut}-\hat{\bS}^{(j)}_{\fut}\|^2_{\text{F}}\big),
    \label{eq:kernel}
\end{equation}
where $\alpha\,{>}\,0$ is a parameter, $\theta_{ij}\,{\in}\,[0,\pi]$ is the un-oriented angle between segments 
$\big(\bS_{\past}(T_{\past}\big),\hat{\bS}^{(i)}_{\fut}(T_{\fut})\big)$ and $\big(\bS_{\past}(T_{\past}),\hat{\bS}^{(j)}_{\fut}(T_{\fut})\big)$, and $\|.\|_{\text{F}}$ denotes the Frobenius norm.

\subsubsection{\textbf{Quality with a layout loss}}

In order to explicitly control the quality of the forecast trajectories, we leverage the physical constraints given by the drivable area: We introduce a loss term, $\mathcal{L}_{\text{layout}}$, to penalize trajectories predicted out of the drivable area. This binary information is part of the environment map $\bM$: we extract it and apply a Chamfer distance transform on it~\cite{Borgefors1984DistanceTI}. The resulting soft map $\bM_{\text{c}}\in[0,1]^{H\times W}$ allows us to define a  
differentiable objective with respect to the coordinates of a given trajectory. Formally, given $\{\hat{\bS}^{(i)}_{\fut}\}_{i=1\cdots N}$ the $N$ future trajectories predicted by our model from an input pair $(\bS_{\past},\bM)$, our layout loss is defined as:
\begin{equation}
    \mathcal{L}_{\text{layout}}\big(\hat{\bS}^{(1:N)}_{\fut};\bM_c\big) = \sum_{n=1}^N \sum_{t=1}^{T_{\fut}} \bM_{\text{c}}\big(\hat{\bS}^{(n)}_{\fut}(t) \big).
    \label{eq:layloss}
\end{equation}




As demonstrated in \cite{bansal2018chauffeurnet} in the context of imitation learning, providing strong learning signals related to driving rules, such as penalizing off-road driving and collisions, does not work if the predictions are trained to match real-world driving recordings which actually do not include off-road examples. As such, when we generate trajectories that maximize their likelihood under the training data, we cannot make use of such a layout loss because it does not generate enough learning signal. Pairing the inclusion of physical constraints with a diversity-generating mechanism via a gating operation allows us to strike a good balance between diversity and quality. 

\section{Experiments}

Given our proposed architecture and training scheme, we conducted experiments aimed at answering the following questions: (1) How does the addition of a DPP-based training scheme improves the overall diversity? (2) How is the quality of the generated diversity impacted by the layout loss? (3) Is the overall accuracy of the model with respect to the ground truth conserved in experiments on a real-world driving dataset?

\subsection{Metrics}
\laura{As real driving datasets cannot provide ground-truth distribution of possible futures, we need tools to measure the diversity of the generated trajectories while ensuring that the single ground-truth future is among predictions.}

\smallskip\noindent\textbf{Diversity and admissibility metrics.~} Measuring diversity is not straightforward since it is not a well-defined concept. 
As a consequence, several diversity metrics have been proposed.  
The ratio of average Final Distance Error (FDE) to min FDE, $\text{rF} = \frac{\text{avgFDE}}{\text{mFDE}}$ \cite{park2020diverse}, is a measure of the spread of the proposed trajectories relative to the ground truth: A high value indicates a high avgFDE, meaning some predictions are far away from the ground truth, and a small mFDE, i.e, one of these predictions is close to the ground truth. 
To measure the spread of the predicted set independently from the ground truth, we use the Average Self Distance (ASD) and Final Self Distance (FSD), introduced in \cite{yuan2019diverse}.
We also use two additional metrics to add a qualitative assessment of the diversity: 
the Drivable Area Occupancy (DAO) \cite{park2020diverse}, which measures the diversity in predictions that are in the drivable area and the Drivable Area Count (DAC) \cite{chang2019argoverse} defined as $\text{DAC} = \frac{N-m}{N}$, where $m$ is the number of predictions that exit the drivable area (DA). 


As discussed in \cite{park2020diverse}, these metrics are complementary: 
rF, ASD and FSD quantify the diversity in terms of mere spread; 
DAC only assesses the admissibility of the trajectories in the set;
DAO captures a mix of diversity and admissibility, by measuring the spread among admissible trajectories only. 
DAO and DAC, being directly related to the drivable area, they provide valuable insights into how the proposed trajectories resemble real trajectories. We were not able to reproduce \cite{park2020diverse}, so we do not report the FSD and ASD metrics as they were not originally evaluated in the paper. This does not impair the results as DAO and rF provide a good diversity assessment.

\smallskip\noindent\textbf{Ground-truth metrics.~} \laura{In addition to diversity metrics, we evaluate the accuracy of our method with respect to the dataset's  unique ground-truth trajectory, traditionally assessed with an Euclidean distance.} Following the existing trajectory forecasting literature \cite{ park2020diverse, phan2020covernet, yuan2019diverse, yuan2020dlow, bansal2018chauffeurnet, rhinehart2018r2p2}, we use the minimum Average Distance Error (mADE) and Final Distance Error (mFDE), computing the error on respectively all the points of the trajectory or only the final one.

\subsection{Experimental setup}

\noindent\textbf{Dataset.~} nuScenes \cite{nuscenes2019} is a real-world driving dataset consisting of around 850 driving scenes of 20 seconds each. These scenarios were recorded in Boston and Singapore, respectively left and right-hand traffic regions, and include complex maneuvers and layouts. Annotations are very detailed and allow the use of ``bird's-eye-view'' (BEV) maps containing information such as drivable area and pedestrian crossings. However, as a real-world dataset, it offers only a single ground-truth future trajectory for each past trajectory, which makes learning multiple-output prediction difficult. 

\noindent\textbf{DIVA setup.~} Our experiments are conducted with the following parameters: The loss balancing coefficient $\lambda$ is set to $0.5$, the latent dimension $d_z$ to $16$ and the past-embedding dimension $d_h$ to $128$. More details can be found in Supplementary A and in our associated code (https://github.com/lcalem/DIVA).

\subsection{Results and discussion}

\noindent\textbf{Results.~} We compare our method to three baselines using a generative backbone: for methods using a cVAE backbone, we compare with \cite{lee2017desire} as a reference for cVAE generative models without any explicit diversification mechanism, and with \cite{yuan2019diverse}, a diversity method also using DPPs. Originally tested on a toy dataset for trajectory prediction, we report here the results of \cite{yuan2019diverse} when tested on the real-world dataset nuScenes.
We also include a comparison with CAM-NF \cite{park2020diverse}, a recent method involving a diversification mechanism built upon a Normalizing Flow (NF) attentional backbone with the whole drivable area as an equiprobable ground-truth distribution for possible futures.
 As for other diversity methods, we measure diversity and admissibility for 3s predictions with 6s of past history.

\begin{table}[h!]
\setlength\tabcolsep{4.5pt}
\begin{tabular}{c c | c | c c c} 
 \toprule
 Model & Backbone & $\text{mADE}$ / $\text{mFDE} \downarrow$ & DAO $\uparrow$ & DAC $\uparrow$ & rF $\uparrow$ \\ [0.5ex] 
 \midrule
DESIRE \cite{lee2017desire} & cVAE & 1.079 / 1.844 & 16.29 & 0.776 & 1.717  \\
L2 DPP \cite{yuan2019diverse}$^*$ & cVAE & 1.148 / 2.272 & 13.31 & \textbf{0.975} & 1.891 \\ 
\name & cVAE & \textbf{0.942} / \textbf{1.449} & \textbf{34.99} & 0.972 & \textbf{4.907} \\ 
 \toprule
 CAM-NF \cite{park2020diverse} & NF-A & 0.639 / 1.171 & 22.62 &	0.918 & 2.558 \\
\bottomrule

\end{tabular}
\caption{\textbf{Prediction assessment on nuScenes}. Evaluation of quality, diversity and admissibility metrics (computed on $N=12$ predictions) for 3s forecast by our best model and cVAE-backbone baselines. 
We also include CAM-NF \cite{park2020diverse} for the sake of completeness, even though it has a different backbone, preventing comparisons. *: Our implementation.}
\label{table:compa}
\end{table}

Results in \autoref{table:compa} indicate that our best model, including the DPP loss with a combined angle and Gaussian 
kernel, has the best performance, improving the diversity both in quantity, as measured with the spread relative to the ground truth (rF), and also in quality.
For completeness, we included in our results the mADE and mFDE metrics which measure the precision of the best prediction compared to the ground truth, although the focus of this work is on diversity. These metrics depend mostly on the generative model used during the initial training, which explains the better precision on these metrics of \cite{park2020diverse} which has a backbone relying on attention mechanisms and normalizing flows. 
We use a cVAE backbone and obtain results similar to those of DESIRE and L2 DPP on these metrics, as expected due to the generative backbone being the same. All diversity metrics (DAO, rF and DAC) show a marked increase 
compared to \cite{park2020diverse} despite the simpler backbone, showing the significance of our contribution on diversity.

\smallskip\noindent\textbf{Ablation study.~} To analyze the contributions of our architecture and losses, we perform an ablation study, the results of which can be seen in \autoref{table:ablation}. As a baseline, we start by training a very simple cVAE backbone with the loss given by \autoref{eq:vae_loss}, and assess its performance on the predicted trajectories decoded from $\bp$, $\bem$ and a $\bz$ component sampled from the Gaussian prior.

\begin{table}[h!]
\setlength\tabcolsep{4.4pt}
\begin{tabular}{l c | c c c c c} 
 \toprule
 Model & $\text{mADE}$/$\text{mFDE}\downarrow$ & DAO & DAC & rF  & ASD  & FSD$\uparrow$ \\ [0.5ex] 
 \midrule
cVAE  & 1.374 / 2.682 & 10.91 & 0.975 & 1.246 & 0.120 & 0.165 \\
\midrule
DSF 1B D & 1.152 / 2.275 & 13.04 & 0.975 & 1.881 & 0.642 & 0.872 \\
DSF 1B L & 1.383 / 2.723 & ~5.81 & 0.977 & 1.046 & 0.048 & 0.058 \\
\midrule
DSF 2B D & 1.018 / 1.594  & 35.08 & 0.917 & 4.948 & 2.319 & 3.033 \\
DSF 2B (D+L)  & 0.942 / 1.449 & 34.99 & 0.972 & 4.907 & 2.142 & 2.842 \\
\bottomrule 
\end{tabular}
\caption{\textbf{Impact of each component}. Evaluation of the contribution of each component to the quality and diversity of $N=12$ predictions over 3s on nuScenes.}
\label{table:ablation}
\end{table}

As expected, the results of the cVAE baseline are relatively mediocre on the quality metrics mADE and mFDE, although consistent with the results of DESIRE, which makes use of a ``rank-and-refine'' module in addition to the cVAE. Low scores of DAO, rF, ASD and FSD are also expected, since the model fails to diversify the predictions and essentially predicts stacked trajectories that go straight. This outcome also explains well the very high DAC measure, as the prediction almost exits the drivable area.
By replacing the Gaussian sampling by a ``weak'' DSF composed of only one branch (`DSF 1B'), improvements on diversity are seen when training the DSF with the diversity loss (`D') but not when training with the layout loss only (`L'); this is expected as the layout loss does not enforce any diversity constraints.
When using our two-branch DSF architecture (`DSF 2B') with an element-wise product to combine $\bz_m$ and $\bz_p$, significant improvements in diversity occur. If the DSF is trained using the diversity loss only (`D'), diversity scores are at their maximum, at the expense of the DAC metric which shows that some (8.3\%) trajectories exit the drivable area as a result of the diversification.
Adding the layout loss (`D+L') improves the quality again to levels comparable to the non-diverse baseline cVAE, at the expense of a slight drop in raw spread as indicated by the decrease in ASD and FSD ($-3.67\%$ and $-5.67\%$ respectively).

\begin{figure*}[b]
    \centering
    \includegraphics[width=0.98\textwidth]{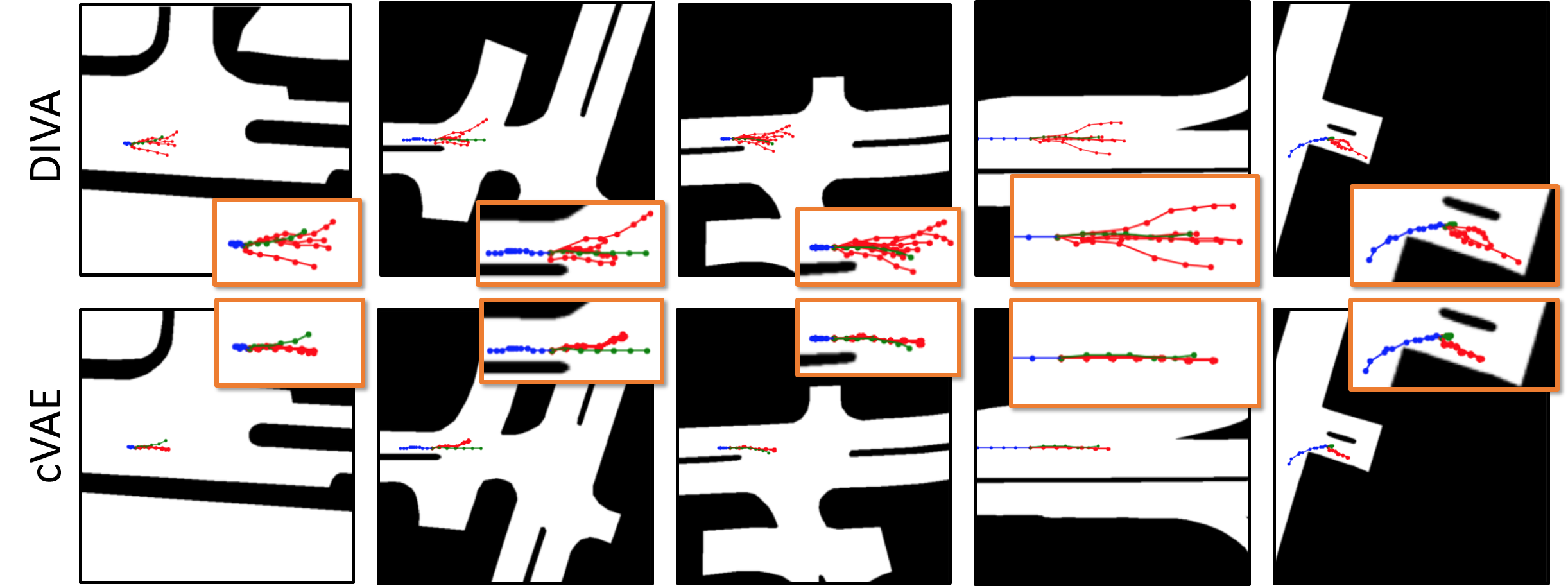}
    \vspace{0pt}
    \caption{\textbf{Qualitative results for various scene layouts in nuScnes}. (Top) Results from proposed DIVA model. (Bottom) Results on the same scenes with a simple cVAE, showing a focus on longitudinal diversity (speed) at the great expense of lateral diversity (direction). In each scene: past and future ground-truth trajectories are in blue and green, resp., while predicted future trajectories are in red (best viewed in color).}
    \label{fig:qu}
\end{figure*}

\smallskip\noindent\textbf{Fusion.} As discussed \autoref{sec:model}, the fusion between layout and diversity encodings $\mathbf{z}_m$ and $\mathbf{z}_p$ is a crucial feature of our model. We compare various fusions in \autoref{table:fusion} and highlight that the best results are obtained by the element-wise product, validating the gating hypothesis. 


\begin{table}[h]
\setlength\tabcolsep{4.4pt}
\addvbuffer[7pt 0pt]{\begin{tabular}{l c | c c c c c} 
 \toprule
 Model & $\text{mADE}$ / $\text{mFDE} \downarrow$ & DAO & DAC & rF & ASD & FSD $\uparrow$ \\ [0.5ex] 
 \midrule
concat & 1.146 / 2.270 & 12.293 & 0.972 & 1.765 & 0.633 & 0.858 \\ 
sum $\bigoplus$ & 1.007 / 1.833 & 28.175 & 0.935 & 3.083 & 1.474 & 1.881\\
product $\bigodot$ & 0.942 / 1.449 & 34.992 & 0.972 & 4.907 & 2.142 & 2.842\\
 \bottomrule
\end{tabular}}
\caption{\textbf{Ablation of the fusion between layout and diversity encodings}. Metrics computed on $N=12$ predictions over 3s in nuScenes, with three ways to combine $\mathbf{z}_m$ and $\mathbf{z}_p$.}
\label{table:fusion}
\end{table}

\subsection{Model analysis} \label{analysis}

\begin{figure}[!h]
    \centering
    \includegraphics[width=0.43\textwidth]{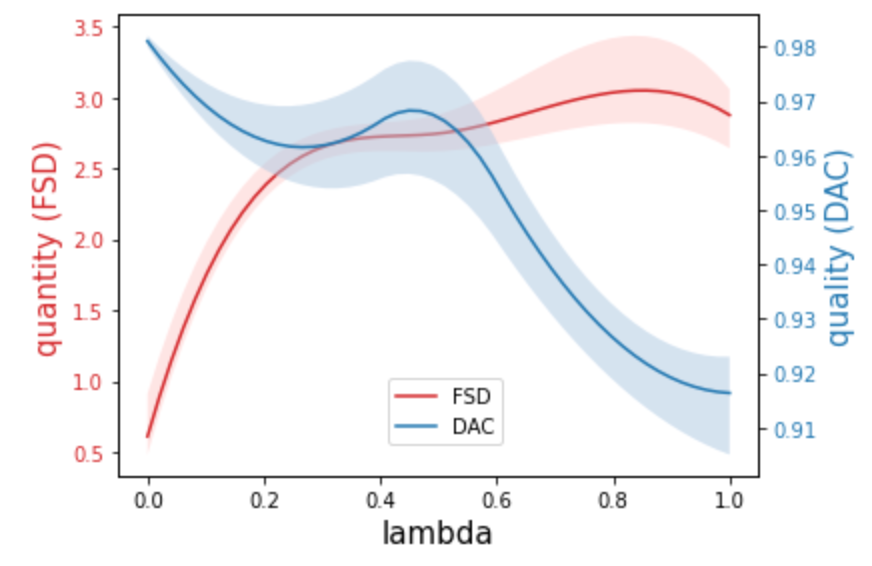}
    \caption{\textbf{Impact of the balance between quality and quantity losses}. Influence of the weighting parameter in the loss $\mathcal{L}_{\text{dsf}} = \lambda \mathcal{L}_{\text{dpp}} + (1 - \lambda) \mathcal{L}_{\text{layout}}$, measured by FSD (red, left axis) and DAC (blue, right axis). For $\lambda=0$, diversity is suppressed and trajectories stay in the drivable area but have low spread; at $\lambda=1$, the diversity is maximal at the expense of admissibility.}
    \label{fig:lambda}
\end{figure}


\noindent\textbf{Loss balancing.~} 
In \autoref{fig:lambda}, we show the effects of varying the balance between the $\mathcal{L}_{\text{layout}}$ and $\mathcal{L}_{\text{past}}$ loss terms in \autoref{eq:dsf_loss}. The axes are chosen to be FSD as a measure of diversity quantity (as it measures the spread of the proposed trajectories) and DAC as a measure of diversity quality (as it measures the percentage of proposed trajectories that stay in the drivable area), to best show the tradeoff between quantity of diversity and quality of this diversity when varying $\lambda$. At one extreme $\lambda = 0$, we suppress the DPP loss entirely, resulting in a very low diversity and a very high quality. The other extreme, $\lambda = 1$, zeroes out the layout loss (although we still have both DSF branches in the architecture), yielding, as expected, results similar to the second row of \autoref{table:ablation}.

\noindent\textbf{Qualitative results.~} In \autoref{fig:qu}, we highlight the diversity improvements from our model on a variety of scenes, including intersections, straight lines and parking. Note how the basic cVAE model exhibits a mode collapse, and how the diversity of our model is influenced by both the past trajectory and the layout of the scene. Additional qualitative results are also available in Supplementary B.

\textcolor{white}{-}
\section{Conclusion}

In this paper, we introduce \name, a multi-output forecasting method for predicting diverse yet admissible trajectories. We \laura{use} a DPP probabilistic model for diversity, and introduce a specific DPP kernel for predicting diverse driving options, \laura{leveraging the variety of settings present in the training data}. The compatibility of the proposed diverse set with the drivable area is controlled by the inclusion of an admissibility loss independent from the underlying generative model. Quantitative and qualitative experiments on real-world dataset nuScenes confirm the benefit on diversity of the proposed architecture and training scheme.






\bibliographystyle{IEEEtran}
\bibliography{IEEEabrv,refs}

%



\end{document}


%
\title{Supplementary Material}


%


\maketitle


%
\IEEEpeerreviewmaketitle

\section{Model details}

\subsection{Bird's-eye-view (BEV) maps} 
The BEV maps represent a 50m by 50m square region in the real world around an agent (25m on each side of the agent, 40m in front of it and 10m behind it). They are encoded as RGB $H\times W$-sized images, where $H = W =224$.  
Following prior work, these BEV maps are constructed by superimposing both static scene elements (drivable area, pedestrian crossings, stop signs, lanes) and dynamic ones (other agents), each category being distinctively color-coded. In addition, for each surrounding agent, previous positions are also reported (with time-stamped color coding).  
            
\subsection{Conditional varational autoencoder (cVAE)} 
The recurrent encoder of the cVAE model is a unidirectional GRU \cite{cho2014gru} with $d_h = 128$ units, so that $\mathbf{h} \in \mathbb{R}^{128}$. 
The CNN encoder for BEV maps is a ResNet18 \cite{he2016deep} with a final layer mapping the output to $d_m = 128$, giving $\mathbf{m} \in \mathbb{R}^{128}$.
Finally, the decoder is composed of a unidirectional GRU with $d_h + d_m + d_z = 128 + 128 + 16 = 272$ hidden units, paired with a fully-connected layer to project this result onto the final output trajectory space of $T_\text{f} = 6$ future steps.

\subsection{Diversity sampling function (DSF)} 
The DSF is made of two identical branches comprised of four fully-connected layers of size 512, with batch normalization \cite{ioffe2015batchnorm} and leaky ReLU activations. 

\subsection{Parameter $\alpha$} 
The definition of kernel $K$ in Eq. 4 includes a parameter $\alpha$ to improve convergence. As long as this parameter is positive, the kernel remains positive semi-definite and the DPP formulation is valid. Setting this parameter to $\alpha = \frac{1}{N^2} \sum_{i=1}^N \sum_{j=1}^N (\theta_{ij} + \| \hat{\bS}^{(i)}_{\fut}-\hat{\bS}^{(j)}_{\fut}\|^2_{\text{F}})$, to calibrate the kernel around the mean of inner kernel values for a particular set $\mathcal{Y}$, works best and is a method often used in kernel methods \cite{shawe2004kernel} and particularly in DPPs \cite{guen2020stripe}.

\section{Additional visualizations}

\begin{figure*}
    \includegraphics[width=1\textwidth]{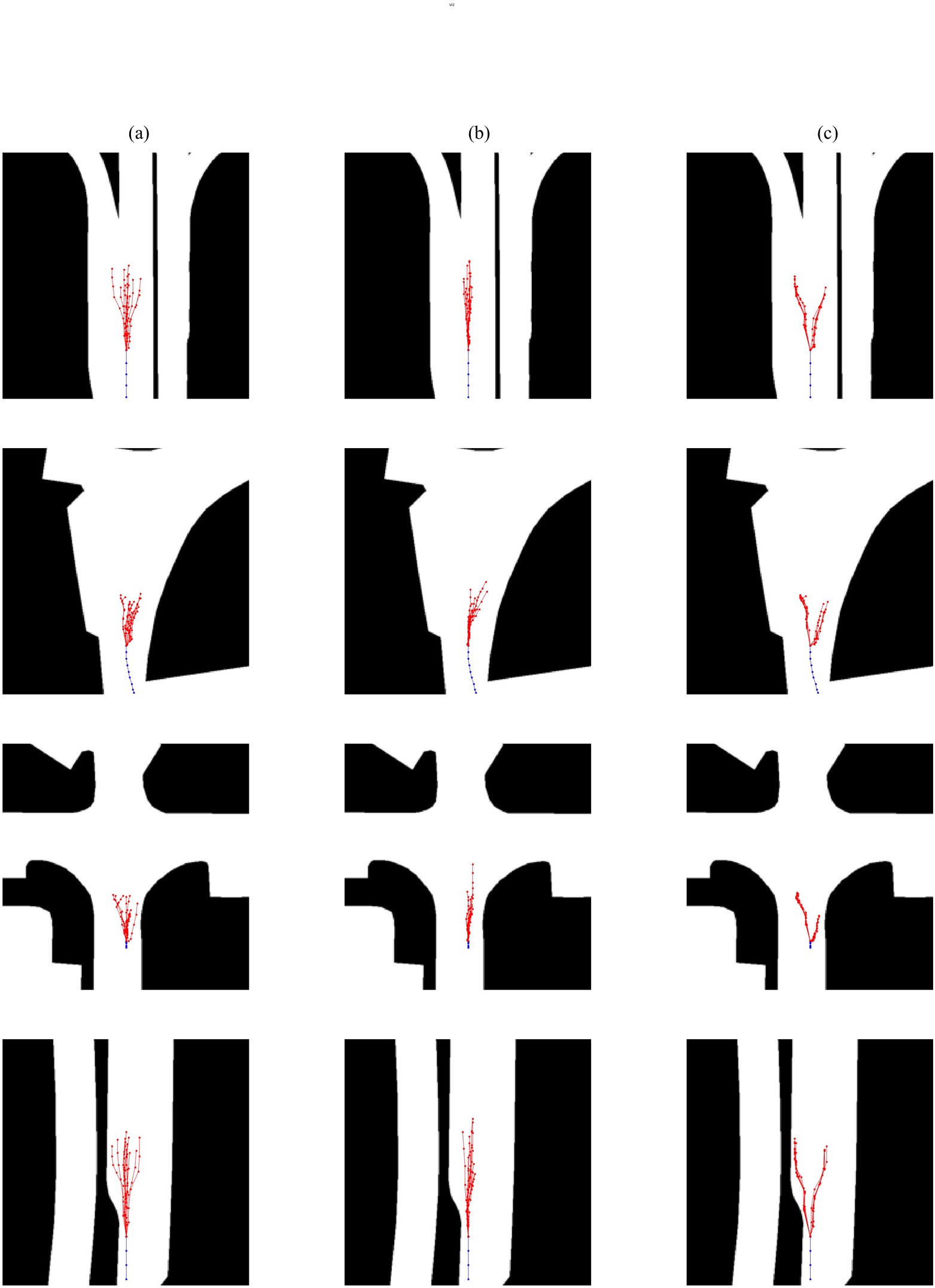}
    \vspace{0pt}
    \caption{\textbf{Effects of different kernels in various layouts}. (a) Our compound angular/distance kernel strikes a good balance between speed and direction sampling; (b) A distance L2 kernel allows speed sampling only; (c) An angular kernel samples two modes of driving direction.}
    \label{fig:kcompa}
\end{figure*}

To illustrate the benefit of our compound angular/distance kernel $K$, we compare its behaviour in \autoref{fig:kcompa}, to the ones of distance-only and angle-only kernels in various situations. 
We see that the distance kernel produces temporally ``dilated'' trajectories (speed variations along a single trajectory) as this is the easiest way to maximize the diversity without exiting the drivable area; The angular kernel exhibits a ``two-mode collapse'' regime as angles are computed pairwise; In contrast, our hybrid kernel samples well both speeds and directions while respecting environment constraints.

\bibliographystyle{IEEEtran}
\bibliography{IEEEabrv,refs}

%

